\documentclass[journal,twoside]{IEEEtran}
\usepackage{cite}
\usepackage{amsmath,amssymb,amsfonts}
\usepackage{graphicx}
\usepackage{booktabs}
\usepackage{multirow}
\usepackage{textcomp}
\usepackage[hidelinks]{hyperref} 
\usepackage{xurl}  
\usepackage{algorithm}
\usepackage{algpseudocode}
\usepackage{caption}
\usepackage{dblfloatfix}

\usepackage{float}

\def\BibTeX{{\rm B\kern-.05em{\sc i\kern-.025em b}\kern-.08em
    T\kern-.1667em\lower.7ex\hbox{E}\kern-.125emX}}
\begin{document}
\title{Quantum CT via Dynamic Interval Encoding and Prior-Balanced QUBO Reconstruction}
\author{ Ao Wang, Yikuang Yuluo, Yujie Liu, Shuangyang Zhong, Yuwen Zhang, Zihao Wang, Fenglin Liu, \IEEEmembership{Senior Member, IEEE}, Andreas Maier, \IEEEmembership{Senior Member, IEEE}, Haijun Yu and Yixing Huang
\thanks{A. Wang and Y.K. Yuluo contributed equally to this work. Corresponding authors: Y.X. Huang and H.J. Yu.}
\thanks{A. Wang, S.Y. Zhong, Y.W. Zhang,  H.J. Yu, and Y.X. Huang are with the Institute of Medical Technology, Peking University Health Science Center, Beijing 100191, China (wao@stu.cqu.edu.cn, 2511210765@stu.pku.edu.cn, 2310117139@stu.pku.bjmu.cn, yuhj@pku.edu.cn, and huangyx@pku.edu.cn).}
\thanks{A. Maier is with the Pattern Recognition Lab (LME), Department of Computer Science, Friedrich-Alexander-Universit\"at Erlangen-N\"urnberg (FAU), Martensstra{\ss}e 3, 91058 Erlangen, Germany (andreas.maier@fau.de).}
\thanks{Y.K. Yuluo, Y.J. Liu, Z.H. Wang, and F.L. Liu are with the Key Lab of Optoelectronic Technology and Systems and the Engineering Research Center of Industrial Computed Tomography Nondestructive Testing, Ministry of Education, Chongqing University, Chongqing 400044, China (yuluoyikuang@stu.cqu.edu.cn, 202408021038t@stu.cqu.edu.cn, wangzh@stu.cqu.edu.cn, and liufl@cqu.edu.cn).}
}

\maketitle
\begin{abstract}

Quadratic unconstrained binary optimization (QUBO)-based quantum computed tomography (CT) casts reconstruction as a binary quadratic problem for quantum annealing and hybrid quantum--classical solvers. For grayscale CT, however, image encoding is constrained by the binary-variable budget: fixed global bit-plane encodings increase QUBO size and coupling complexity as gray-level precision improves, whereas low-bit encodings introduce quantization error. We propose a QUBO-based grayscale CT reconstruction framework that combines dynamic interval encoding with prior-balanced optimization. Each refinement round encodes active pixels only within local gray-level intervals around the current estimate, and a boundary-hit-guided update rule adaptively switches between search expansion and local refinement. To improve optimization stability, the method balances projection-domain data consistency and an edge-preserving quadratic prior before forming the final QUBO. Sparse-view and limited-angle fan-beam CT experiments show that the proposed method recovers structures and gray-level distributions more faithfully than the evaluated analytic, iterative, variational, and representation-based baselines. Expressivity analysis and ablation studies further indicate that the improvement mainly arises from effective gray-level representation through dynamic local encoding and more stable data-fidelity--prior coupling. Experiments on the D-Wave hybrid binary quadratic model (BQM) solver further demonstrate that the formulation is executable on a hardware-backed hybrid quantum--classical backend.
\end{abstract}


\section{Introduction}
\label{sec:introduction}

\IEEEPARstart{Q}{uadratic} unconstrained binary optimization (QUBO)-based quantum computed tomography (CT) casts tomographic reconstruction as a binary quadratic problem for quantum annealing and hybrid quantum--classical solvers \cite{kadowaki1998quantum,lucas2014ising}. In this framework, image intensities are encoded by binary variables, and projection consistency is expressed as a quadratic objective. Unlike continuous-variable reconstruction, QUBO-based reconstruction makes the image representation part of the optimization problem itself: improving gray-level precision usually requires more binary variables and increases the size, coupling density, and coefficient range of the QUBO. Therefore, for grayscale CT, reconstruction quality is limited not only by projection data and regularization, but also by the binary-variable budget available for image encoding.

\par Existing QUBO-based tomographic studies have shown that CT reconstruction can be cast as a binary optimization problem solvable by annealing or hybrid solvers \cite{nau2023exploring,haga2024quantum,jun2023highly,dremel2025utilizing,jun2025quantum}. These studies also indicate that image encoding is a key factor governing scalability and reconstruction accuracy. For binary images, low-integer-valued images, or material labels, a small number of candidate states may be sufficient to describe pixel values. Grayscale CT, however, contains continuously varying attenuation coefficients and subtle structural transitions, requiring higher effective gray-level expressivity. A direct fixed global bit-plane encoding increases the number of variables, quadratic couplings, and coefficient dynamic range as the bit depth grows, whereas a low-bit encoding introduces substantial quantization error.

\par We formulate this trade-off as the binary-variable budget problem in grayscale QUBO reconstruction. The binary-variable budget denotes the number of binary variables assigned to image encoding in a single QUBO optimization round. Under a fixed global encoding, this budget is mainly determined by the image size and the number of bits per pixel, so improving gray-level precision typically requires enlarging the QUBO instance directly. The basic idea of this work is to replace a global high-bit representation with a local low-bit representation updated across reconstruction rounds: each round uses only a small number of binary variables to represent local gray-level candidates around the current estimate, and the candidate intervals are iteratively updated to improve effective gray-level resolution.

\par Based on this idea, we propose a QUBO-based grayscale CT reconstruction framework that combines dynamic interval encoding with prior-balanced optimization. Dynamic interval encoding constructs local gray-level candidate sets around the current estimate of each active pixel, allowing each QUBO round to use low-bit variables while improving effective expressivity through interval updates across rounds. A boundary-hit-guided update rule detects insufficient local interval coverage and adaptively switches between search expansion and local refinement. To improve optimization stability, we further combine projection-domain data consistency with an edge-preserving quadratic prior and balance the two objective terms before forming the final QUBO.

\par The main contributions of this work are threefold. First, we identify the binary-variable budget as a representation bottleneck in grayscale QUBO CT reconstruction and introduce dynamic interval encoding to improve effective gray-level expressivity without increasing the per-round bit depth. Second, we develop a boundary-hit-guided interval update mechanism and a prior-balanced QUBO objective, enabling more stable joint optimization of data consistency and an edge-preserving prior. Third, we validate the method on sparse-view and limited-angle fan-beam CT tasks, and further demonstrate in the sparse-view experiments that the formulation can be executed on a D-Wave hybrid binary quadratic model solver.

\section{Related Work}

\subsection{CT Reconstruction Algorithms}

\par CT reconstruction is commonly based on a forward model that maps an attenuation image to projection data, and existing algorithms can be broadly grouped into analytic, iterative, and prior-driven methods. Analytic methods such as filtered backprojection (FBP) \cite{kak1988principles} and the Feldkamp--Davis--Kress (FDK) algorithm \cite{feldkamp1984practical} are computationally efficient and remain important baselines when angular sampling is sufficient, but they are prone to artifacts when the data are incomplete or angular coverage is limited. Algebraic and iterative methods, including algebraic reconstruction technique (ART) \cite{gordon1970algebraic}, simultaneous ART (SART) \cite{andersen1984simultaneous}, and conjugate gradient for normal equations (CGN) \cite{hestenes1952methods,bjorck1979accelerated}, improve reconstruction stability by repeatedly enforcing projection-domain data consistency and reduce the dependence on analytic inversion conditions. These methods form the conventional algorithmic basis of CT reconstruction and provide classical references for the baselines evaluated in this work.

\par Beyond conventional iterative frameworks, regularized, learning-based, and representation-based methods further improve CT reconstruction by introducing image priors or alternative parameterizations. Total variation (TV) \cite{rudin1992nonlinear} and compressed-sensing methods \cite{donoho2006compressed} exploit gradient sparsity for edge-preserving reconstruction, leading to representative algorithms such as SART-TV \cite{sidky2008image} and TV-regularized primal-dual hybrid gradient (TV-PDHG) \cite{chambolle2011first}. Deep learning methods learn reconstruction mappings, iterative updates, or implicit priors from training data or individual images, including convolutional neural network (CNN)-based inverse-problem networks \cite{jin2017deep}, framelet-based sparse-view CT networks \cite{han2018framing}, learned primal-dual reconstruction \cite{adler2018learned}, projection-domain learning for limited-angle CT \cite{wurfl2018deep}, LEARN (learned experts' assessment-based reconstruction network) \cite{chen2018learn}, and deep image prior \cite{ulyanov2018deep}. Recent implicit neural, hash-grid, and Gaussian representations further show that the image representation itself can substantially affect reconstruction quality and parameter efficiency \cite{tian2025unsupervised,li20253dgr,wu2025discretized}. These methods provide an important context for this work: reconstruction performance depends not only on the objective function, but also on how the image is parameterized. Unlike these continuous-parameter representations, this work focuses on binary image representations compatible with QUBO optimization.

\subsection{QUBO-Based Quantum CT Reconstruction}

\par QUBO-based quantum CT differs from quantum-state image reconstruction in that it seeks binary optimization solutions that can be decoded into classical CT images, rather than quantum-state outputs with nontrivial pixel-wise readout \cite{kiani2020quantum}. In this class of methods, pixels or image states are encoded by binary variables, and projection-domain errors or related reconstruction objectives are transformed into quadratic forms compatible with quantum annealing, adiabatic computation, or hybrid quantum--classical solvers. Existing studies have demonstrated the feasibility of this direction, including hybrid adiabatic reconstruction for small binary or integer-valued images \cite{nau2023exploring} and low-qubit real-valued approximation with two qubits per pixel \cite{haga2024quantum}. These works also reveal a common limitation: as image size, the number of pixel states, or the bit range increases, the number of QUBO variables, coupling complexity, and solution difficulty can grow rapidly.

\par Subsequent studies have further expanded the objective design, data types, and encoding strategies of QUBO-based tomographic reconstruction. Existing work has investigated sinogram-discrepancy QUBO objectives \cite{jun2023highly}, compressed-sensing or TV-related constraints \cite{ryou2026quantum}, quantum-annealing reconstruction on simulated and measured CT data \cite{dremel2025utilizing}, and material-coefficient-based candidate reduction for joint reconstruction and segmentation \cite{jun2025quantum}. These studies advance QUBO-based CT from different perspectives, but most existing formulations still rely on fixed bit-plane encodings, a small number of candidate states, material labels, or low-qubit approximations. For grayscale CT, such encodings remain limited in simultaneously maintaining a compact QUBO size and sufficient gray-level expressivity. In contrast, this work does not assign each pixel a fixed global high-bit code; instead, it uses local dynamic interval encoding updated across reconstruction rounds and further improves the numerical coupling between data consistency and image priors through objective-term balancing.

\section{Methods}

\begin{figure*}[!t]
\centerline{\includegraphics[width=1.5\columnwidth]{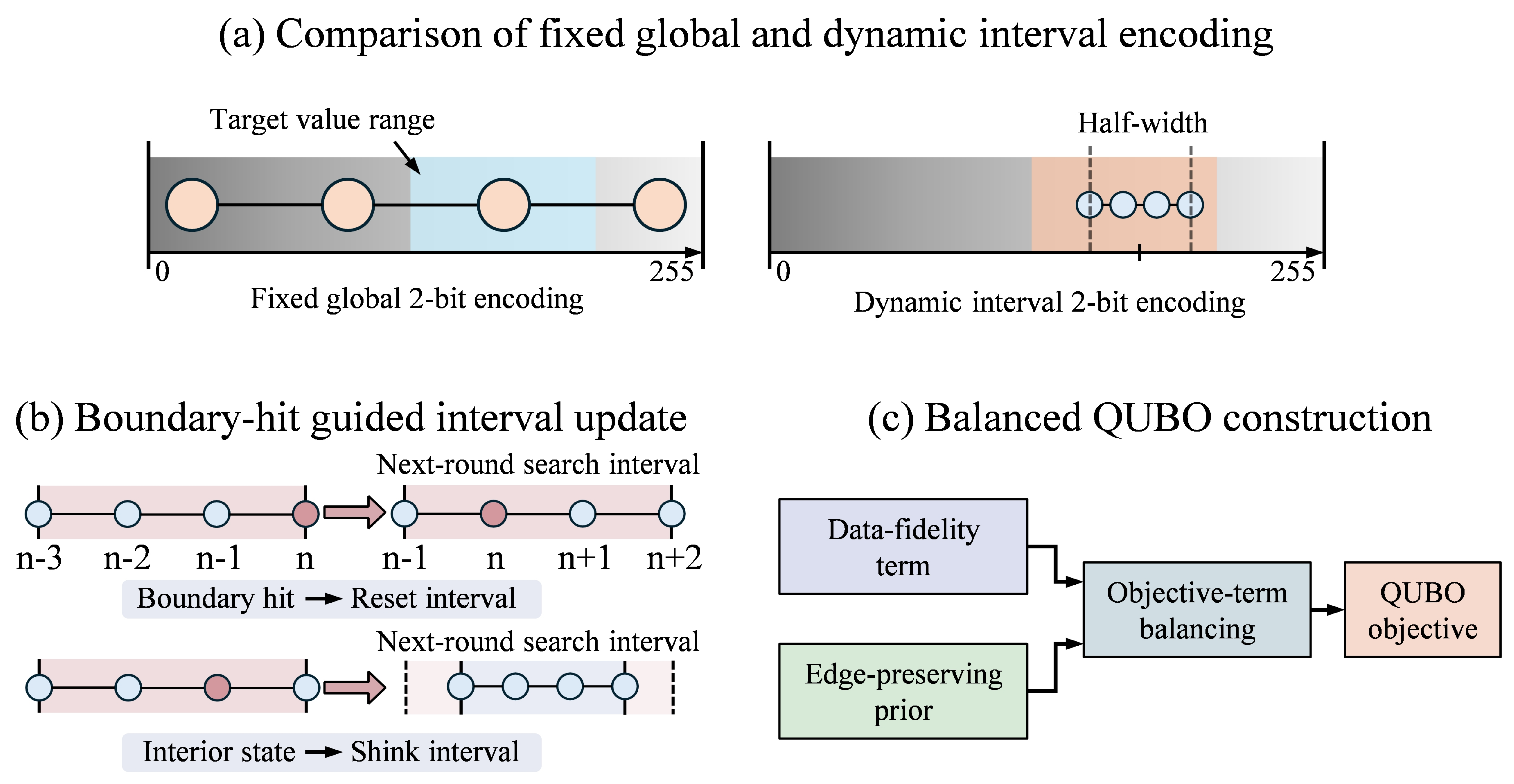}}
\caption{Overview of the proposed framework: (a) dynamic interval encoding versus fixed global encoding under a 2-bit budget, (b) boundary-hit-guided interval update, and (c) balanced QUBO construction with an edge-preserving prior.}
\label{framework}
\end{figure*}

\par Fig.~\ref{framework} summarizes the proposed reconstruction framework and illustrates its three core components in sequence: dynamic interval encoding, boundary-hit-guided interval updating, and balanced QUBO construction with an edge-preserving prior.

\subsection{Problem Formulation} 
\par Let $\boldsymbol{x} \in \mathbb{R}^N$ denote the unknown CT image, $\boldsymbol{y} \in \mathbb{R}^M$ the measured projection data, and $\boldsymbol{A} \in \mathbb{R}^{M \times N}$ the system matrix. The forward model is

\begin{equation}
\boldsymbol{y} = \boldsymbol{A}\boldsymbol{x} + \boldsymbol{\varepsilon},
\end{equation}
where $\boldsymbol{\varepsilon}$ denotes noise and modeling error.
\par In classical CT reconstruction, $\boldsymbol{x}$ is typically estimated by solving

\begin{equation}
\min_{\boldsymbol{x}} \mathcal{D}(\boldsymbol{A}\boldsymbol{x}, \boldsymbol{y}) + \lambda \mathcal{R}(\boldsymbol{x}),
\end{equation}
where $\mathcal{D}(\cdot, \cdot)$ denotes a data-fidelity term and $\mathcal{R}(\boldsymbol{x})$ denotes a prior or regularization term.
\par In QUBO-based quantum CT, reconstruction does not optimize the continuous image directly. Instead, image intensities are represented through a binary decision vector $\boldsymbol{z} \in \{0, 1\}^K$ and a discrete image mapping

\begin{equation}
\boldsymbol{x} = \Phi(\boldsymbol{z}),
\end{equation}
where $\Phi$ maps binary variables to image intensities and $K$ is the total number of binary variables. Thus, $K$ specifies the binary-variable budget of the current QUBO instance; for a uniform $B$-bit encoding over an active set $\Omega_{\mathrm{act}}$, this budget is $K = B|\Omega_{\mathrm{act}}|$. The corresponding reconstruction objective can be written as

\begin{equation}
\min_{\boldsymbol{z} \in \{0, 1\}^K} \mathcal{D}(\boldsymbol{A}\Phi(\boldsymbol{z}), \boldsymbol{y}) + \lambda \mathcal{R}(\Phi(\boldsymbol{z})).
\end{equation}
When the data-fidelity and prior terms can be reduced to quadratic functions of $\boldsymbol{z}$, the reconstruction can be written in QUBO form as

\begin{equation}
\min_{\boldsymbol{z} \in \{0, 1\}^K} \boldsymbol{z}^{\top}\boldsymbol{Q}\boldsymbol{z} + \boldsymbol{q}^{\top}\boldsymbol{z} + c,
\end{equation}
where $\boldsymbol{Q}$ and $\boldsymbol{q}$ are the quadratic matrix and linear coefficient vector, respectively, and $c$ is a constant.
\par Under this formulation, the central design question is the choice of $\Phi$. A global fixed bit-plane encoding is straightforward, but it becomes inefficient as image size and bit depth increase, especially when additional prior terms are introduced. We therefore seek a progressive local discrete representation that preserves image expressivity while keeping the binary model compact. This motivates the dynamic interval encoding described next.

\subsection{Dynamic Interval Encoding} 
\label{subsec:encoding} 

\par Instead of assigning a fixed global bit-plane representation to every pixel, we encode each active pixel using a local candidate set around the current estimate. As illustrated in Fig.~\ref{framework}(a), the proposed method replaces fixed global low-bit encoding with a dynamic local interval representation under the same 2-bit budget. Let $c_p^{(t)}$ and $r_p^{(t)}$ denote the center and half-range of pixel $p$ at refinement iteration $t$, respectively. For each active pixel $p \in \Omega_{\mathrm{act}}$, the interval is initialized by setting

\begin{equation}
c_p^{(0)} = x_p^{(0)}, \qquad r_p^{(0)} = r_{\mathrm{init}},
\end{equation}

where $x_p^{(0)}$ is the initial image value at pixel $p$, and $r_{\mathrm{init}}$ is a user-defined initial search half-width measured in the same gray-value units as the image intensity. The nominal local search interval at round $t$ is
\begin{equation}
\mathcal{I}_p^{(t)} = [ \, c_p^{(t)} - r_p^{(t)}, \, c_p^{(t)} + r_p^{(t)} \, ].
\end{equation}

\par For each active pixel, we use a low-bit local representation within $\mathcal{I}_p^{(t)}$. Specifically, the pixel value is expressed as
\begin{equation}
x_p^{(t)} = b_p^{(t)} + \sum_{k=1}^B w_{p,k}^{(t)} z_{p,k}^{(t)}, \quad z_{p,k}^{(t)} \in \{0, 1\},
\end{equation}
where $B$ is the number of local bits, $b_p^{(t)}$ is the local base value, and $w_{p,k}^{(t)}$ are the corresponding bit weights. In this work, we use $B = 2$, which gives four candidate values per pixel at each round.

\par The local base value is defined as
\begin{equation}
b_p^{(t)} = \operatorname{clip}\left(c_p^{(t)} - r_p^{(t)},\, 0,\, x_{\max} - 2r_p^{(t)}\right),
\end{equation}
where $\operatorname{clip}(a,l,u)=\min\{\max(a,l),u\}$ constrains $a$ to the interval $[l,u]$. This operation ensures that the lower endpoint of the local candidate range is nonnegative and that the upper endpoint does not exceed $x_{\max}$. Pixels outside the active set are excluded from encoding and are fixed to zero; therefore, binary variables are assigned only to pixels that are allowed to change.
\par The spacing between adjacent local candidate values is then given by
\begin{equation}
\Delta_p^{(t)} = \frac{2r_p^{(t)}}{2^B - 1},
\end{equation}
which reduces to $\Delta_p^{(t)} = 2r_p^{(t)}/3$ in the $B=2$ case used in this work. The bit weights are set to $w_{p,k}^{(t)} = 2^{k-1}\Delta_p^{(t)}$ for $k=1,\ldots,B$, so that the candidate set becomes
\begin{equation}
\left\{
b_p^{(t)},\,
b_p^{(t)}+\Delta_p^{(t)},\,
b_p^{(t)}+2\Delta_p^{(t)},\,
b_p^{(t)}+3\Delta_p^{(t)}
\right\}.
\end{equation}
This construction ensures that the local interval is covered by four uniformly spaced discrete states while remaining within the valid image range.
\par Relative to global fixed-range encoding, the key difference is that the proposed representation is adaptive and progressive: each refinement iteration explores only a small neighborhood around the current estimate, and the local interval is updated across iterations. This design preserves local grayscale expressivity while keeping the binary model compact, which is advantageous for QUBO construction and hardware-constrained optimization.

\subsection{Boundary-Hit-Guided Interval Update} 
\par After solving the binary optimization problem at refinement iteration $t$, each active pixel is assigned one discrete state within its local interval. Let $s_p^{(t)}$ denote the decoded state index of pixel $p$. For the 2-bit case,
\begin{equation}
s_p^{(t)} = z_{p,1}^{(t)} + 2z_{p,2}^{(t)}, \quad s_p^{(t)} \in \{0, 1, 2, 3\}.
\end{equation}

\par The interval center is then updated to the decoded value obtained from the current QUBO solution, denoted by $\hat{x}_p^{(t)}$, i.e.
\begin{equation}
c_p^{(t+1)} = \hat{x}_p^{(t)}, \quad p \in \Omega_{\text{act}},
\end{equation}
where $\Omega_{\text{act}}$ denotes the active pixel set. Pixels outside the active set remain fixed.

\par The interval width is updated according to whether the decoded state hits the boundary of the current local interval. We define a boundary hit as
\begin{equation}
s_p^{(t)} \in \{0, 2^B - 1\}.
\end{equation}

\par If a boundary hit occurs, the local interval half-width is reset to the initial search half-width used at $t=0$,
\begin{equation}
r_p^{(t+1)} = r_{\mathrm{init}}.
\end{equation}

\par Otherwise, the interval is contracted for local refinement,
\begin{equation}
r_p^{(t+1)} = \max(r_{\text{min}}, \rho(r_p^{(t)})),
\end{equation}
where $\rho(\cdot)$ denotes a shrink operator. In this implementation, we use a multiplicative shrinkage rule,
\begin{equation}
\rho(r) = \operatorname{round}(\alpha r),
\end{equation}
with $\alpha \in (0, 1)$, and enforce the minimum search half-width through $r_{\min}$.

\par Fig.~\ref{framework}(b) illustrates the interval-adaptation rule: a boundary hit resets the next-iteration search interval, whereas an interior state triggers interval shrinkage. This rule provides a simple exploration-refinement mechanism. A boundary hit indicates insufficient local interval coverage, whereas an interior state suggests a more stable local region. The method therefore expands the search conservatively when needed and otherwise contracts the interval to improve local precision.

\subsection{Balanced QUBO Construction} 
\par At each refinement iteration, the image is represented by the local binary variables introduced in Section~\ref{subsec:encoding}. Substituting this representation into the reconstruction objective yields a binary optimization problem with a data-fidelity term and a prior term.

\par Let $\boldsymbol{z}$ denote the binary variable vector at the current iteration. The per-round energy can be written as
\begin{equation}
E(\boldsymbol{z}) =  \lambda_{\mathrm{data}}E_{\mathrm{data}}(\boldsymbol{z}) + \lambda_{\mathrm{prior}} E_{\mathrm{prior}}(\boldsymbol{z}).
\end{equation}

\par For the data-fidelity term, we first separate the current image estimate into a fixed base component and a binary residual component. This yields
\begin{equation}
E_{\mathrm{data}}(\boldsymbol{z}) = \| \boldsymbol{y} - \boldsymbol{A}\boldsymbol{x}_{\mathrm{base}} - \tilde{\boldsymbol{A}}\boldsymbol{z} \|_2^2,
\end{equation}
where $\boldsymbol{x}_{\mathrm{base}}$ is the base image induced by the local interval encoding and $\tilde{\boldsymbol{A}}$ is the effective system matrix associated with the binary residual variables. Expanding the quadratic form gives
\begin{equation}
E_{\mathrm{data}}(\boldsymbol{z}) = \boldsymbol{z}^{\top}\boldsymbol{Q}_{\mathrm{data}}\boldsymbol{z} + \boldsymbol{q}_{\mathrm{data}}^{\top}\boldsymbol{z} + c_{\mathrm{data}}.
\end{equation}

\par To preserve local smoothness while avoiding excessive penalization across structural transitions, we introduce an edge-preserving prior in quadratic form,
\begin{equation}
E_{\mathrm{prior}}(\boldsymbol{z}) = \sum_{(u,v) \in \mathcal{N}} w_{uv} \left( x_u(\boldsymbol{z}) - x_v(\boldsymbol{z}) \right)^2,
\end{equation}
where $\mathcal{N}$ is the neighborhood set and $w_{uv}$ is an adaptive edge weight. In this work, $w_{uv}$ is computed from a reference image $\boldsymbol{x}^{\mathrm{ref}}$, which is updated from the current reconstruction at each round,
\begin{equation}
w_{uv} = \frac{1}{\sqrt{(x_u^{\mathrm{ref}} - x_v^{\mathrm{ref}})^2 + \epsilon^2}},
\end{equation}
with $\epsilon > 0$ for numerical stability. This leads to another quadratic binary form,
\begin{equation}
E_{\mathrm{prior}}(\boldsymbol{z}) = \boldsymbol{z}^{\top}\boldsymbol{Q}_{\mathrm{prior}}\boldsymbol{z} + \boldsymbol{q}_{\mathrm{prior}}^{\top}\boldsymbol{z} + c_{\mathrm{prior}}.
\end{equation}

\par Directly summing these two terms is often unstable because the coefficient scales of the data and prior parts can differ substantially. We therefore balance them before combination. Let $\mathcal{C}_{\mathrm{data}}$ and $\mathcal{C}_{\mathrm{prior}}$ denote the sets of absolute nonzero coefficients in the data and prior QUBO terms, including both quadratic and linear coefficients. With $P_{\eta}(\cdot)$ denoting the $\eta$-th percentile and $\tau_{\mathrm{data}}$ and $\tau_{\mathrm{prior}}$ denoting prescribed target magnitudes, the normalization factors are defined as
\begin{equation}
s_{\mathrm{data}} = \frac{\tau_{\mathrm{data}}}{P_{\eta}(\mathcal{C}_{\mathrm{data}})+\delta}, \quad
s_{\mathrm{prior}} = \frac{\tau_{\mathrm{prior}}}{P_{\eta}(\mathcal{C}_{\mathrm{prior}})+\delta},
\end{equation}
where $\delta$ is a small constant for numerical stability. The final QUBO is written as
\begin{equation}
E(\boldsymbol{z}) = \boldsymbol{z}^{\top}\boldsymbol{Q}\boldsymbol{z} + \boldsymbol{q}^{\top}\boldsymbol{z} + c,
\end{equation}
with
\begin{equation}
\boldsymbol{Q} = \lambda_{\mathrm{data}} s_{\mathrm{data}} \boldsymbol{Q}_{\mathrm{data}} + \lambda_{\mathrm{prior}} s_{\mathrm{prior}} \boldsymbol{Q}_{\mathrm{prior}},
\end{equation}
\begin{equation}
\boldsymbol{q} = \lambda_{\mathrm{data}} s_{\mathrm{data}} \boldsymbol{q}_{\mathrm{data}} + \lambda_{\mathrm{prior}} s_{\mathrm{prior}} \boldsymbol{q}_{\mathrm{prior}}.
\end{equation}

\par This percentile-based normalization avoids letting a few extreme coefficients dominate the scaling and makes the relative effects of $\lambda_{\mathrm{data}}$ and $\lambda_{\mathrm{prior}}$ more interpretable across iterations and acquisition settings. As shown in Fig.~\ref{framework}(c), the data-fidelity term and the edge-preserving prior are combined through objective-term balancing to form the final QUBO objective.

\subsection{Overall Reconstruction Framework} 

\par Algorithm~\ref{alg:framework} summarizes the proposed round-wise workflow. Starting from an initial image, the method identifies active pixels, initializes local intervals, builds a balanced QUBO with data fidelity and an edge-preserving prior, solves the binary problem, and updates interval centers and widths according to the boundary-hit rule.

\begingroup
\footnotesize
\setlength{\abovedisplayskip}{2pt}
\setlength{\belowdisplayskip}{2pt}
\noindent\rule{\linewidth}{0.8pt}\vspace{-0.5\baselineskip}
\captionof{algorithm}{Progressive quantum CT reconstruction with dynamic interval encoding.}
\label{alg:framework}
\vspace{-0.5\baselineskip}\noindent\rule{\linewidth}{0.4pt}

\begin{algorithmic}[1]
\Require $\boldsymbol{y}$, $\boldsymbol{A}$, $\boldsymbol{x}^{(0)}$, $\Omega_{\mathrm{act}}$, $B$, $r_{\mathrm{init}}$, $r_{\min}$, $x_{\max}$, $T$, $\alpha$, $\lambda_{\mathrm{data}}$, $\lambda_{\mathrm{prior}}$
\Ensure Reconstructed image $\boldsymbol{x}$

\State $\boldsymbol{x} \gets \boldsymbol{x}^{(0)}$
\ForAll{$p \in \Omega_{\mathrm{act}}$}
    \State $(c_p, r_p) \gets (x_p^{(0)}, r_{\mathrm{init}})$
\EndFor

\For{$t = 0,1,\dots,T-1$}
    \Statex \textbf{Dynamic local interval encoding}
    \ForAll{$p \in \Omega_{\mathrm{act}}$}
        \State $b_p \gets \operatorname{clip}(c_p-r_p,\,0,\,x_{\max}-2r_p)$, \quad
        $\Delta_p \gets \dfrac{2r_p}{2^B-1}$
        \State Build local candidates and residual $\boldsymbol{z}$
    \EndFor

    \Statex \textbf{Balanced QUBO construction}
    \State Build $E_{\mathrm{data}}(\boldsymbol{z})$ and $E_{\mathrm{prior}}(\boldsymbol{z})$ with adaptive edge weights
    \State Compute $s_{\mathrm{data}}$ and $s_{\mathrm{prior}}$; form $E(\boldsymbol{z})=\boldsymbol{z}^{\top}\boldsymbol{Q}\boldsymbol{z}+\boldsymbol{q}^{\top}\boldsymbol{z}+c$
    \Statex \quad $\boldsymbol{Q}=\lambda_{\mathrm{data}}s_{\mathrm{data}}\boldsymbol{Q}_{\mathrm{data}}+\lambda_{\mathrm{prior}}s_{\mathrm{prior}}\boldsymbol{Q}_{\mathrm{prior}}$
    \Statex \quad $\boldsymbol{q}=\lambda_{\mathrm{data}}s_{\mathrm{data}}\boldsymbol{q}_{\mathrm{data}}+\lambda_{\mathrm{prior}}s_{\mathrm{prior}}\boldsymbol{q}_{\mathrm{prior}}$

    \Statex \textbf{Solve and interval update}
    \State Solve the QUBO, decode $s_p$, and reconstruct $\boldsymbol{x}_{\mathrm{new}}$
    \ForAll{$p \in \Omega_{\mathrm{act}}$}
        \State $c_p \gets x_{\mathrm{new},p}$
        \State $r_p \gets r_{\mathrm{init}}$ if $s_p\in\{0,2^B-1\}$
        \State otherwise, $r_p \gets \max(r_{\min},\operatorname{round}(\alpha r_p))$
    \EndFor
    \State $\boldsymbol{x} \gets \boldsymbol{x}_{\mathrm{new}}$
\EndFor

\State \Return $\boldsymbol{x}$
\end{algorithmic}

\par\vspace{0.2\baselineskip}
\noindent\rule{\linewidth}{0.8pt}
\endgroup

\section{Experiments and Results}
\label{sec:experiments}

\subsection{Experimental Protocol}

\subsubsection{Dataset and preprocessing}
\par
\begin{figure}
    \centering
    \includegraphics[width=0.9\linewidth]{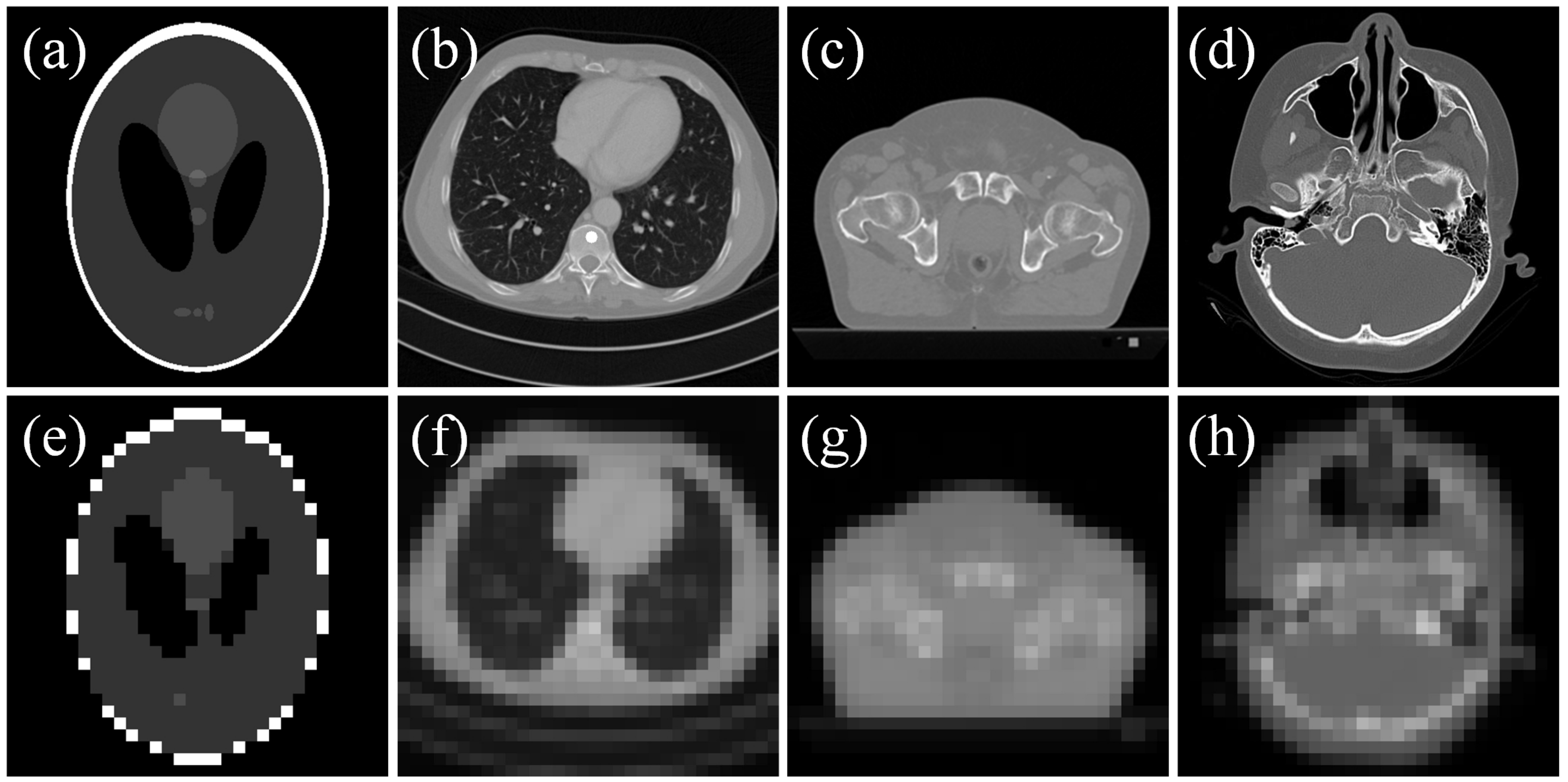}
    \caption{Reference images $512 \times 512$ and their $32 \times 32$ versions used to generate projections and compute metrics: Shepp--Logan (a,e), American Association of Physicists in Medicine (AAPM) CT slices (b,f) and (c,g), and Kaggle head CT (d,h).}
    \label{fig:test_images}
\end{figure}

 To evaluate the proposed reconstruction framework under different structural characteristics, we used four representative images, including one Shepp--Logan phantom, two clinical CT slices from the American Association of Physicists in Medicine (AAPM) low-dose dataset \cite{moen2021lowdose}, and one head CT slice from the Kaggle dataset \emph{Computed Tomography (CT) of the Brain} \cite{kagglebrainct}. The first row of Fig.~\ref{fig:test_images} shows the original reference images, and the second row shows the resized inputs used for reconstruction. All experiments were conducted on \(32 \times 32\) images to keep the QUBO instances small enough for proof-of-concept evaluation while preserving recognizable anatomical or structural variation. Before projection generation, all images were resized using bilinear interpolation, and their grayscale values were mapped to the range \([0,255]\).

\subsubsection{Projection geometry and acquisition protocols}
\par Projection generation and reconstruction were implemented with the ASTRA toolbox \cite{van2016fast} under a 2D fan-beam geometry. The source-to-object distance (SOD) and object-to-detector distance (ODD) were both set to 100 mm, corresponding to a source-to-detector distance (SDD) of 200 mm. The detector consisted of 100 bins with a 1\,mm bin spacing. Unless otherwise stated, no measurement noise was added, so the experiments focus on reconstruction difficulty caused by angular incompleteness and discrete QUBO modeling rather than stochastic measurement corruption.

\par The main sparse-view benchmark used uniform views over \(0^\circ\!-\!360^\circ\). Images 1--2 were reconstructed from 9 views, whereas Images 3--4 were reconstructed from 18 views because 9 views caused severe degradation for all methods on the latter two cases. All methods were evaluated under matched geometry and view settings for each image. The average reported in Table~\ref{tab:quantitative_results} should therefore be interpreted as a cross-case summary rather than as the result of a single shared acquisition protocol.

\par In addition to the full-range sparse-view benchmark, we conducted a supplementary limited-angle robustness experiment on a representative phantom. Four angular spans, \(60^\circ\), \(90^\circ\), \(120^\circ\), and \(150^\circ\), were considered. For each angular span, 18 projection views were uniformly sampled within the available angular range. All other reconstruction parameters were kept the same as in the sparse-view experiments.

\subsubsection{Implementation details of the proposed method}

\par For the proposed method, reconstruction used 50 refinement iterations. At each iteration, each active pixel was encoded by a 2-bit local interval, yielding four candidate gray levels within the current search interval. The maximum gray value was \(x_{\max}=255\). The initial and minimum interval half-widths were \(r_{\mathrm{init}}=5\) and \(r_{\min}=1\) on the \([0,255]\) gray-value scale, and interval shrinkage used a multiplicative factor of 0.7. Thus, before valid-range clipping, the initial local search interval spans at most \(2r_{\mathrm{init}}=10\) gray levels around the initial pixel estimate. The adaptive-edge-weight reference image was initialized from a 200-iteration SART reconstruction. The active pixel set $\Omega_{\mathrm{act}}$ comprised pixels with strictly positive values in this initial SART reconstruction, serving as a coarse foreground mask to focus binary variables on likely object-support regions. The objective weights were $\lambda_{\mathrm{data}}=1.0$ and $\lambda_{\mathrm{prior}}=4.0$.

\par We implemented two solver variants: a local simulated-annealing solver, denoted SA (Ours), and the D-Wave hybrid binary quadratic model (BQM) solver \cite{dwave2020hybrid}, denoted Hybrid (Ours). SA used \texttt{num\_reads = 5} and \texttt{num\_sweeps = 10000}, whereas Hybrid used the solver-determined default \texttt{time\_limit} provided by the D-Wave service. Both variants followed the same 50-round dynamic-interval refinement schedule.

\par Runtime was reported per QUBO solver call. SA time was recorded as \texttt{sa\_time\_sec}, excluding QUBO construction, interval updating, image decoding, metric computation, and file operations, and was averaged over 10 repeated calls for each sparse-view case. Hybrid time was the solver-side \texttt{run\_time} returned by D-Wave. These values characterize solver cost and backend executability rather than end-to-end acceleration or quantum speedup.

\subsubsection{Baselines, evaluation metrics, and hardware}

\par We compared the proposed method with analytic, iterative, variational, and representation-based baselines, including FBP, SART, CGN, SART-TV, TV-PDHG, a Gaussian-based representation baseline, and self-prior embedding neural representation (Spener) \cite{tian2025unsupervised}. FBP used the Ram--Lak filter. SART and CGN were run for 200 and 100 iterations, respectively. SART-TV used 10 outer iterations, each consisting of 20 SART updates followed by one TV-denoising step. TV-PDHG used a TV weight of 0.1 and 150 iterations. The Gaussian-based method used 4000 Gaussian points and was trained for 2000 iterations with \(\lambda_{\mathrm{tv}}=0.05\) and \(\lambda_{\mathrm{dssim}}=0.25\). Spener used a hash-grid encoder with 5 levels and 8 features per level, followed by a 2-layer multi-layer perceptron (MLP) with 64 hidden units per layer, and was trained for 3000 epochs with $\lambda=2.5$. The Gaussian-based method and Spener were evaluated under matched projection geometry, with the remaining settings kept fixed across experiments.

\par Reconstruction quality was evaluated using peak signal-to-noise ratio (PSNR), structural similarity index measure (SSIM) \cite{wang2004image}, and normalized root mean square error (NRMSE). Higher PSNR and SSIM indicate better reconstruction quality, whereas lower NRMSE indicates smaller pixel-wise error. All PSNR, SSIM, and NRMSE values were computed over the full \(32 \times 32\) image.

\par All classical baselines and the SA variant were executed on a local workstation with an Intel Core i7-14700KF CPU, 64 GB RAM, and an NVIDIA GeForce RTX 5060 Ti GPU (16 GB). The Hybrid variant was executed on the D-Wave platform with the solver \texttt{hybrid\_binary\_quadratic\_model\_version2p} and a binary quadratic model (BQM) formulation.

\subsection{Sparse-View Reconstruction and Hybrid Execution}

\subsubsection{Main sparse-view results}

\begin{figure*}[!t]
\centerline{\includegraphics[width=1.95\columnwidth]{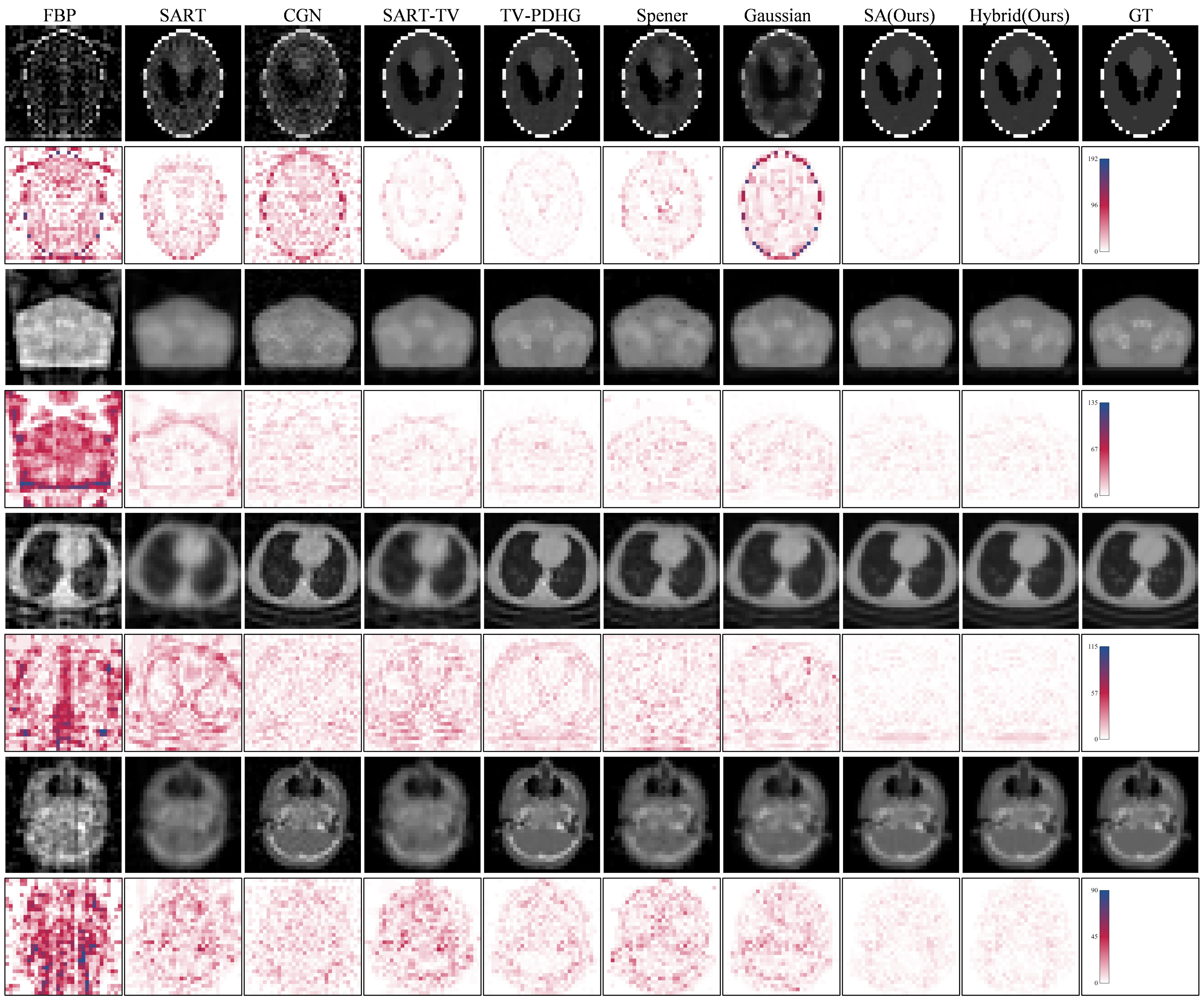}}
\caption{Sparse-view reconstruction results. For each test image, the top row shows reconstructions and the bottom row shows absolute error maps. The proposed SA and Hybrid variants reduce streak artifacts and residual errors.}
\label{fig:main_results}
\end{figure*}

\begin{table*}[!t]
\centering
\caption{Quantitative sparse-view comparison. Images 1--2 use 9 views and Images 3--4 use 18 views; the reported average is computed across the four image cases. For SA (Ours), the parenthesized row reports the standard deviation over 10 repeated SA runs and is not included in the ranking. Bold and underlined entries indicate the best and second-best mean results, respectively.}
\label{tab:quantitative_results}
\setlength{\tabcolsep}{3.5pt}
\renewcommand{\arraystretch}{1.10}
\resizebox{\textwidth}{!}{
\begin{tabular}{lccc ccc ccc ccc ccc}
\toprule
\multirow{2}{*}{Method} 
& \multicolumn{3}{c}{Image 1} 
& \multicolumn{3}{c}{Image 2} 
& \multicolumn{3}{c}{Image 3} 
& \multicolumn{3}{c}{Image 4} 
& \multicolumn{3}{c}{Avg. across four cases} \\
\cmidrule(lr){2-4}\cmidrule(lr){5-7}\cmidrule(lr){8-10}\cmidrule(lr){11-13}\cmidrule(lr){14-16}
& PSNR$\uparrow$ & SSIM$\uparrow$ & NRMSE$\downarrow$
& PSNR$\uparrow$ & SSIM$\uparrow$ & NRMSE$\downarrow$
& PSNR$\uparrow$ & SSIM$\uparrow$ & NRMSE$\downarrow$
& PSNR$\uparrow$ & SSIM$\uparrow$ & NRMSE$\downarrow$
& PSNR$\uparrow$ & SSIM$\uparrow$ & NRMSE$\downarrow$ \\
\midrule
Hybrid (Ours)
& \underline{45.10} & \textbf{0.9819} & \underline{0.0056}
& \textbf{39.14} & \underline{0.9600} & \textbf{0.0110}
& \textbf{37.93} & \textbf{0.9819} & \textbf{0.0127}
& \textbf{40.36} & \textbf{0.9823} & \textbf{0.0096}
& \textbf{40.63} & \underline{0.9765} & \textbf{0.0097} \\

SA (Ours)
& \textbf{45.13} & \underline{0.9819} & \textbf{0.0055}
& \underline{39.01} & \textbf{0.9622} & \underline{0.0112}
& \underline{37.66} & \underline{0.9812} & \underline{0.0131}
& \underline{39.89} & \underline{0.9812} & \underline{0.0101}
& \underline{40.42} & \textbf{0.9766} & \underline{0.0100} \\

SA (Ours) std.
& $(0.12)$ & $(0.0002)$ & $(0.0001)$
& $(0.16)$ & $(0.0010)$ & $(0.0002)$
& $(0.19)$ & $(0.0004)$ & $(0.0004)$
& $(0.23)$ & $(0.0011)$ & $(0.0009)$
& $(0.18)$ & $(0.0007)$ & $(0.0004)$ \\
\midrule

Gaussian
& 18.91 & 0.7834 & 0.1134
& 34.26 & 0.9452 & 0.0194
& 30.54 & 0.9477 & 0.0297
& 31.86 & 0.9421 & 0.0255
& 28.89 & 0.9046 & 0.0470 \\

Spener
& 28.90 & 0.8995 & 0.0359
& 32.64 & 0.9095 & 0.0233
& 29.29 & 0.8722 & 0.0343
& 30.26 & 0.9071 & 0.0307
& 30.27 & 0.8971 & 0.0311 \\

TV-PDHG
& 35.84 & 0.9377 & 0.0161
& 34.93 & 0.9353 & 0.0179
& 30.30 & 0.9292 & 0.0306
& 33.13 & 0.9324 & 0.0221
& 33.55 & 0.9336 & 0.0217 \\

SART-TV
& 31.40 & 0.9160 & 0.0269
& 34.18 & 0.9235 & 0.0195
& 27.04 & 0.8873 & 0.0445
& 28.58 & 0.8950 & 0.0372
& 30.30 & 0.9055 & 0.0320 \\

CGN
& 20.02 & 0.5643 & 0.0998
& 31.27 & 0.8034 & 0.0273
& 29.77 & 0.8926 & 0.0325
& 31.45 & 0.8501 & 0.0268
& 28.13 & 0.7776 & 0.0466 \\

SART
& 25.25 & 0.8535 & 0.0546
& 28.81 & 0.7615 & 0.0363
& 23.43 & 0.8113 & 0.0674
& 27.75 & 0.8122 & 0.0410
& 26.31 & 0.8096 & 0.0498 \\

FBP
& 16.30 & 0.4822 & 0.1532
& 15.38 & 0.5866 & 0.1702
& 17.75 & 0.7089 & 0.1295
& 19.12 & 0.7077 & 0.1107
& 17.14 & 0.6213 & 0.1409 \\
\bottomrule
\end{tabular}}
\end{table*}

\par Fig.~\ref{fig:main_results} and Table~\ref{tab:quantitative_results} show that, under sparse-view acquisition, the analytic, iterative, variational, and representation-based baselines suffer from different degrees of streak artifacts, contrast bias, or residual errors, whereas the two proposed variants produce structures closer to the ground truth with lower error maps. Quantitatively, Hybrid (Ours) achieves the highest average PSNR and lowest average NRMSE, while SA (Ours) achieves the highest average SSIM, with closely matched results between the two solvers. Compared with the strongest conventional baseline, TV-PDHG, Hybrid improves the average PSNR from 33.55 dB to 40.63 dB and reduces the average NRMSE from 0.0217 to 0.0097.

\par Table~\ref{tab:quantitative_results} also reports the repeated-run stability of SA (Ours). Across 10 SA runs, the standard deviations remain small relative to inter-method differences, indicating good numerical stability in the present setting. The closely matched SA and Hybrid reconstructions under the same formulation suggest that the performance gain mainly comes from dynamic interval encoding and prior-balanced QUBO construction rather than solver-specific behavior. Hybrid further confirms end-to-end execution on a hardware-backed hybrid quantum--classical solver, with reconstruction quality comparable to local SA.

\subsubsection{Execution on a hardware-backed hybrid solver}

\par The Hybrid variant was executed on the D-Wave solver \texttt{hybrid\_binary\_quadratic\_model\_version2p}. As summarized in Table~\ref{tab:hybrid_runtime}, the representative Hybrid runs required 4.160 s per-round QUBO solve on average, with an average quantum processing unit (QPU) access time of 0.076 s and an average QPU ratio of 1.84\%. The local SA solver required 4.952 s per solve on average across the same sparse-view cases. These results document practical solver cost and backend compatibility rather than end-to-end acceleration or quantum speedup.

\begin{table}[t]
\centering
\caption{Solver timing for the sparse-view experiments. Times are reported per-round QUBO solve; SA values are reported as mean $\pm$ standard deviation over 10 solver calls.}
\label{tab:hybrid_runtime}
\setlength{\tabcolsep}{3.4pt}
\renewcommand{\arraystretch}{1.08}
\resizebox{\columnwidth}{!}{
\begin{tabular}{lcccc}
\toprule
Case & Hybrid runtime (s/solve) & QPU access (s/solve) & QPU ratio (\%) & SA time (s/solve) \\
\midrule
Image 1 & 3.465 & 0.104 & 3.00 & $4.814 \pm 0.174$ \\
Image 2 & 3.793 & 0.098 & 2.58 & $4.217 \pm 0.087$ \\
Image 3 & 5.312 & 0.052 & 0.98 & $6.614 \pm 0.702$ \\
Image 4 & 4.068 & 0.052 & 1.28 & $4.162 \pm 1.235$ \\
\midrule
Average & 4.160 & 0.076 & 1.84 & 4.952 \\
\bottomrule
\end{tabular}}
\end{table}

\subsection{Limited-Angle Reconstruction}

\par
\begin{figure}[!t]
\centerline{\includegraphics[width=0.95\columnwidth]{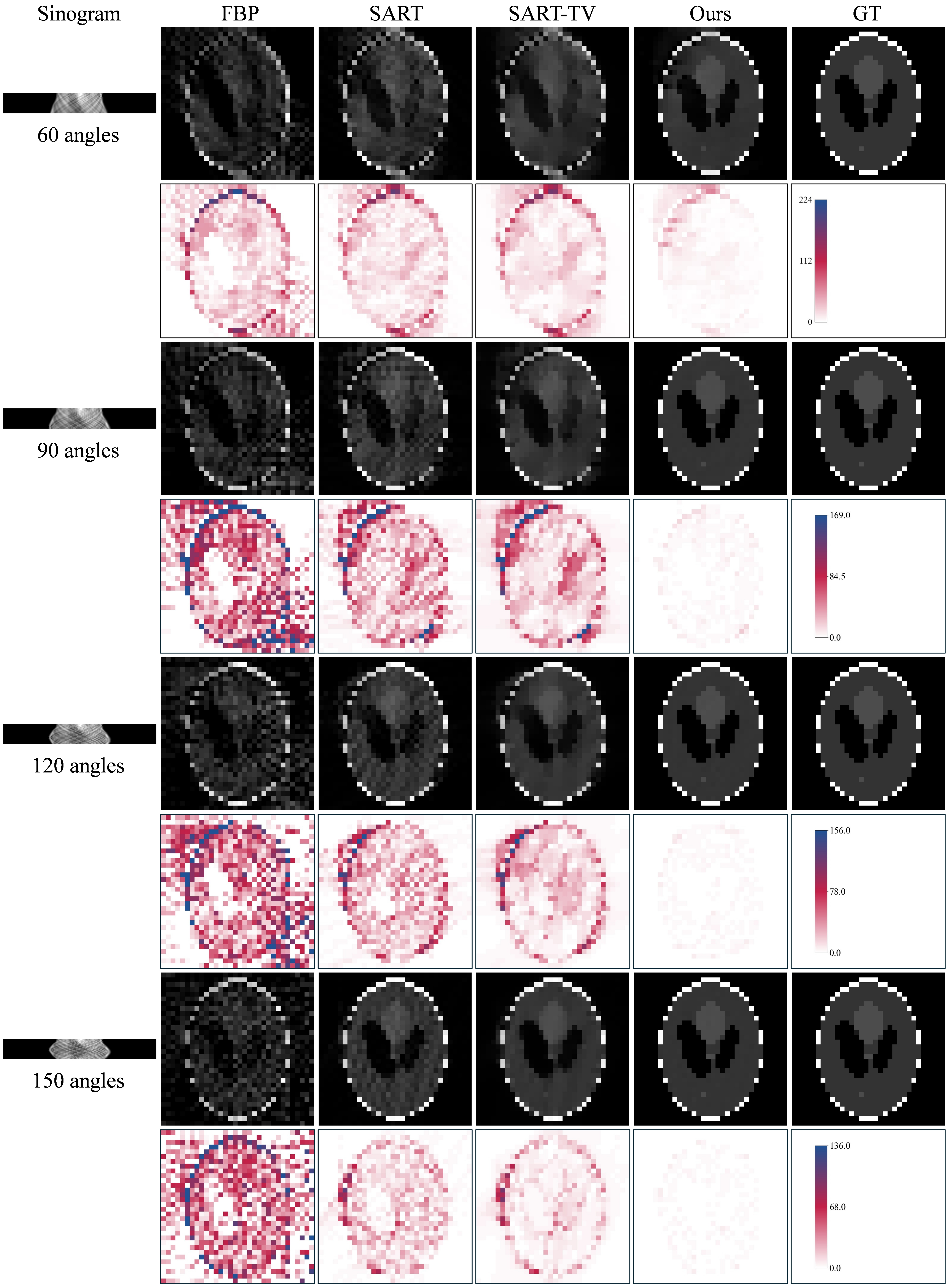}}
\caption{Limited-angle reconstruction on the phantom with 18 views over $60^\circ$, $90^\circ$, $120^\circ$, and $150^\circ$ angular spans. Row labels indicate angular spans; each span shows the sinogram, reconstructions, ground truth, and absolute error maps.}
\label{fig:limitedangle}
\end{figure}

 As a supplementary robustness experiment, we evaluated limited-angle acquisition on a representative phantom. Four angular spans, $60^\circ$, $90^\circ$, $120^\circ$, and $150^\circ$, were considered, with 18 uniformly sampled views in each case. The row labels in Fig.~\ref{fig:limitedangle} denote angular spans rather than the number of projection views. Fig.~\ref{fig:limitedangle} shows that the proposed method better preserves the phantom structure and reduces residual errors.

\par Quantitatively, reconstruction quality improves as the angular span increases from $60^\circ$ to $150^\circ$, suggesting that dynamic local QUBO refinement can exploit additional angular information. This single-phantom test is therefore used as supplementary robustness evidence, not as a comprehensive limited-angle benchmark.

\begin{table}[t]
\centering
\caption{Limited-angle quantitative results for the phantom. Each span uses 18 views; Bold and underlined entries indicate the best and second-best results, respectively.}
\label{tab:limited_angle_results}
\scriptsize
\renewcommand{\arraystretch}{1.08}
\setlength{\tabcolsep}{3.5pt}
\begin{tabular*}{0.92\columnwidth}{@{\extracolsep{\fill}}clcccc@{}}
\toprule
Range & Metric & FBP & SART & SART-TV & SA (Ours) \\
\midrule
\multirow{3}{*}{$60^\circ$}
& PSNR$\uparrow$ & 16.83 & 19.64 & \underline{19.79} & \textbf{29.76} \\
& SSIM$\uparrow$ & 0.6454 & 0.8003 & \underline{0.8110} & \textbf{0.9790} \\
& NRMSE$\downarrow$ & 0.1441 & 0.1042 & \underline{0.1025} & \textbf{0.0325} \\
\midrule
\multirow{3}{*}{$90^\circ$}
& PSNR$\uparrow$ & 17.97 & 22.81 & \underline{23.48} & \textbf{47.80} \\
& SSIM$\uparrow$ & 0.6804 & 0.8495 & \underline{0.8774} & \textbf{0.9995} \\
& NRMSE$\downarrow$ & 0.1264 & 0.0723 & \underline{0.0670} & \textbf{0.0041} \\
\midrule
\multirow{3}{*}{$120^\circ$}
& PSNR$\uparrow$ & 18.71 & 25.39 & \underline{26.47} & \textbf{51.93} \\
& SSIM$\uparrow$ & 0.6886 & 0.9035 & \underline{0.9320} & \textbf{0.9997} \\
& NRMSE$\downarrow$ & 0.1160 & 0.0538 & \underline{0.0475} & \textbf{0.0025} \\
\midrule
\multirow{3}{*}{$150^\circ$}
& PSNR$\uparrow$ & 19.28 & 29.10 & \underline{31.05} & \textbf{53.95} \\
& SSIM$\uparrow$ & 0.6920 & 0.9559 & \underline{0.9769} & \textbf{0.9998} \\
& NRMSE$\downarrow$ & 0.1087 & 0.0351 & \underline{0.0280} & \textbf{0.0020} \\
\bottomrule
\end{tabular*}
\end{table}

\subsection{Formulation Analysis and Sensitivity}

\subsubsection{Encoding expressivity}

\par
\begin{figure}[!t]
\centerline{\includegraphics[width=0.95\columnwidth]{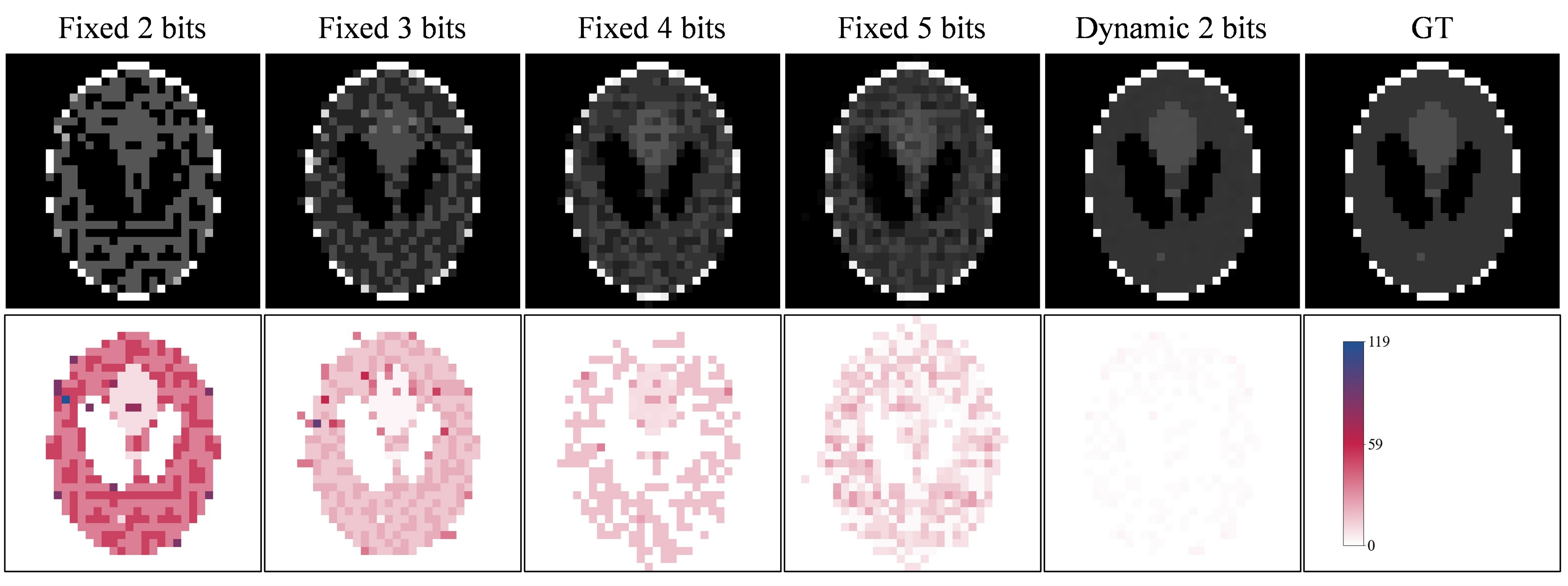}}
\caption{Fixed global encoding versus dynamic interval encoding. Columns show fixed 2--5-bit results, dynamic 2-bit result, and ground truth; error maps are shown below.}
\label{fig:graylevel}
\end{figure}

 We compared fixed global encoding with dynamic interval encoding to analyze grayscale expressivity under a constrained binary-variable budget. Fig.~\ref{fig:graylevel} and Table~\ref{tab:fixed_dynamic_bits} show that increasing the fixed global bit width reduces quantization artifacts, but even fixed 5-bit encoding still exhibits structural distortion and larger residual errors. In contrast, the proposed dynamic 2-bit encoding produces a reconstruction closer to the ground truth.

\begin{table}[!t]
\centering
\caption{Gray-level expressivity comparison for Fig.~\ref{fig:graylevel}. Dynamic 2-bit encoding outperforms fixed global 2--5-bit encodings.}
\label{tab:fixed_dynamic_bits}
\setlength{\tabcolsep}{4pt}
\renewcommand{\arraystretch}{1.12}
\small
\begin{tabular*}{\columnwidth}{@{\extracolsep{\fill}} l c c c c}
\toprule
Strategy & Bit-width & PSNR$\uparrow$ & SSIM$\uparrow$ & NRMSE$\downarrow$ \\
\midrule
Dynamic interval (Ours)& 2-bit & \textbf{51.88} & \textbf{0.9981} & \textbf{0.0025} \\
\midrule
\multirow{4}{*}{Fixed global}
& 2-bit & 19.85 & 0.7879 & 0.1017 \\
& 3-bit & 26.07 & 0.8948 & 0.0497 \\
& 4-bit & 29.94 & \underline{0.9248} & 0.0319 \\
& 5-bit & \underline{30.72} & 0.9144 & \underline{0.0291} \\
\bottomrule
\end{tabular*}
\end{table}

\par Although the proposed strategy uses only 2 local bits at each round, it achieves 51.88 dB PSNR, 0.9981 SSIM, and 0.0025 NRMSE in this representative comparison, compared with 30.72 dB PSNR, 0.9144 SSIM, and 0.0291 NRMSE for fixed 5-bit encoding.

\begin{figure}[!t]
\centerline{\includegraphics[width=1\columnwidth]{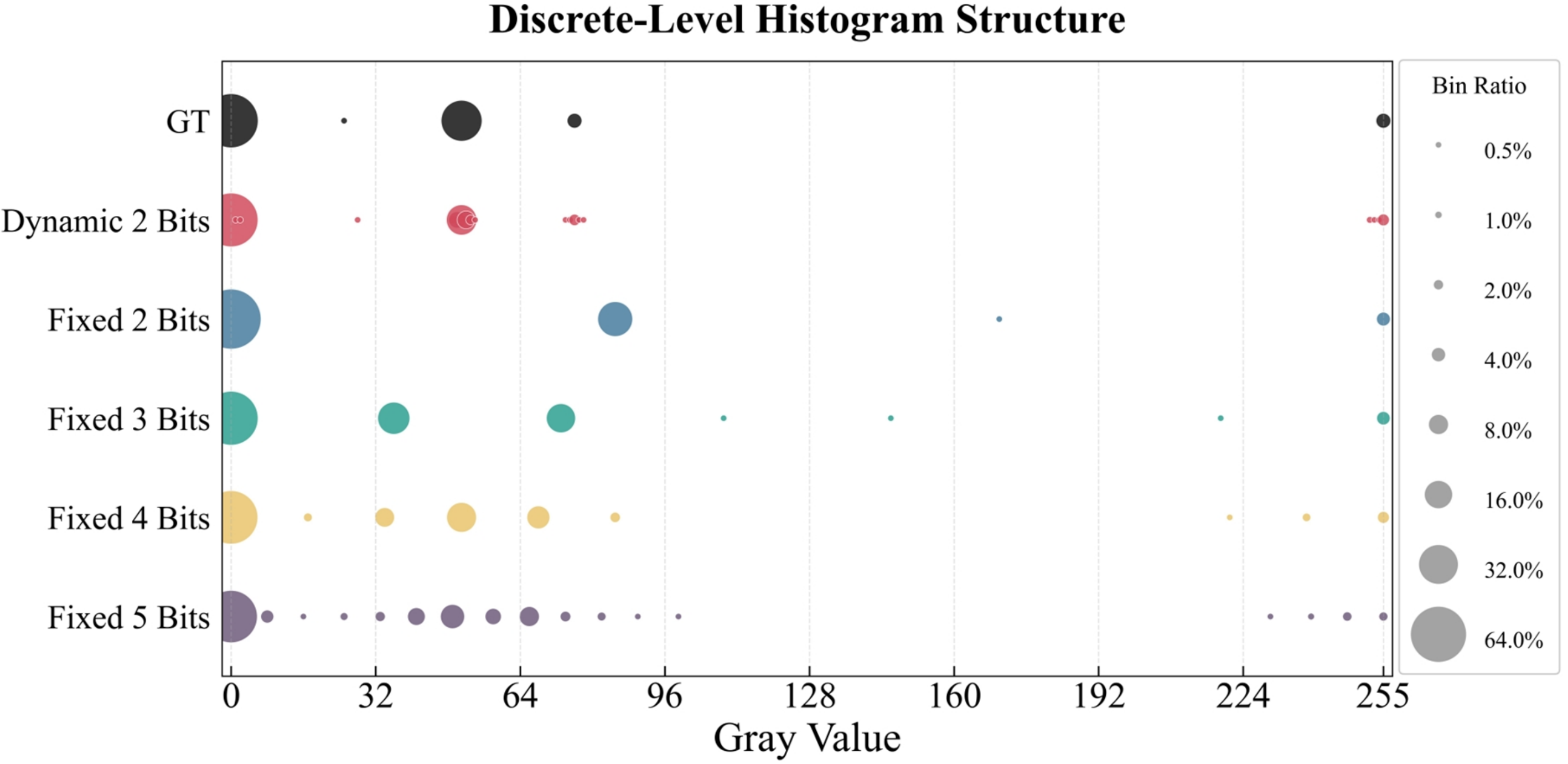}}
\caption{Gray-level occupancy distributions. Bubble size indicates the pixel proportion at each gray level; dynamic 2-bit encoding better matches the ground-truth distribution.}
\label{fig:histogram}
\end{figure}

\par Fig.~\ref{fig:histogram} further explains this behavior from the gray-level occupancy distribution. Fixed-bit strategies concentrate pixels on a small set of globally shared levels, whereas dynamic 2-bit encoding updates local intervals across rounds and produces a denser distribution closer to the ground-truth histogram. These results support the central design goal: improving effective grayscale expressivity without increasing the per-round QUBO bit depth.

\subsubsection{Component ablation}

\par
\begin{figure}[!t]
\centerline{\includegraphics[width=0.7\columnwidth]{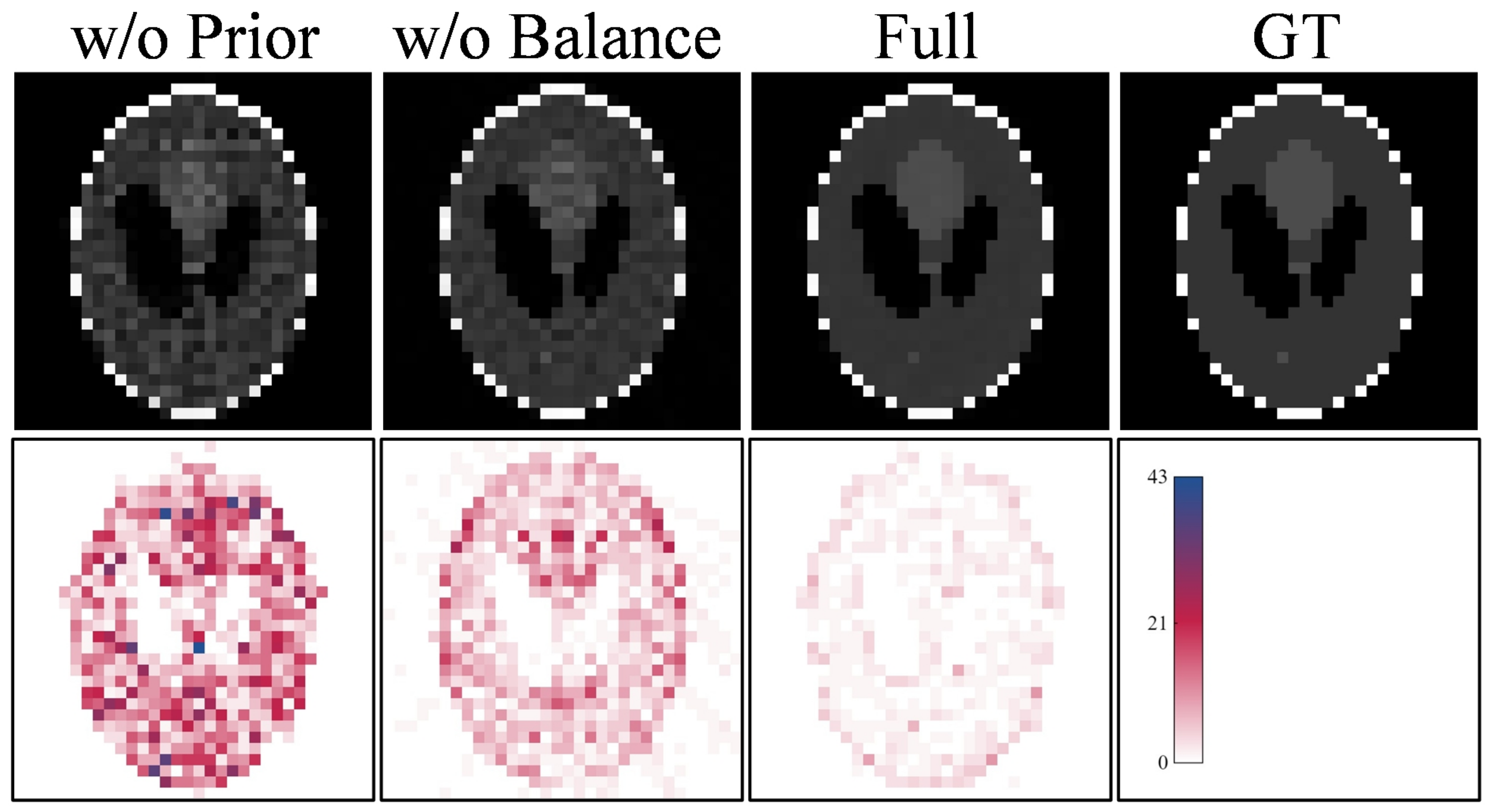}}
\caption{Component ablation. Columns show no prior, no balancing, full model, and ground truth (GT); error maps are shown below. The full model gives the lowest residual error.}
\label{TVorNo}
\end{figure}

 Fig.~\ref{TVorNo} and Table~\ref{tab:ablation_components} evaluate the roles of the edge-preserving prior and objective-term balancing. Removing the prior leads to pronounced artifact contamination and degraded structural continuity, indicating that data fidelity alone is insufficient to stabilize sparse-view reconstruction. Retaining the prior but removing balancing improves the result relative to the no-prior variant, but residual errors remain, suggesting that scale mismatch between the data-fidelity and prior terms can affect joint optimization.

\begin{table}[!t]
\centering
\caption{Quantitative ablation results for Fig.~\ref{TVorNo}. Removing either the prior or term balancing degrades reconstruction quality.}
\label{tab:ablation_components}
\setlength{\tabcolsep}{4pt}
\renewcommand{\arraystretch}{1.12}
\small
\begin{tabular*}{\columnwidth}{@{\extracolsep{\fill}} l c c c}
\toprule
Configuration & PSNR$\uparrow$ & SSIM$\uparrow$ & NRMSE$\downarrow$ \\
\midrule
No prior
& 29.04 & 0.8976 & 0.0353 \\

No balance
& \underline{34.00} & \underline{0.9287} & \underline{0.0200} \\

Full model
& \textbf{45.13} & \textbf{0.9819} & \textbf{0.0055} \\
\bottomrule
\end{tabular*}
\end{table}

\par The full model achieves the highest PSNR/SSIM and lowest NRMSE, indicating that the prior mainly provides structural regularization, while balancing improves the stability of data-fidelity--prior coupling. Relative to the full model, removing balancing lowers PSNR from 45.13\,dB to 34.00\,dB and raises NRMSE from 0.0055 to 0.0200; removing the prior further degrades performance to 29.04\,dB PSNR and 0.0353 NRMSE.

\subsubsection{Parameter sensitivity}
\par
\begin{figure}[!t]
\centerline{\includegraphics[width=0.95\columnwidth]{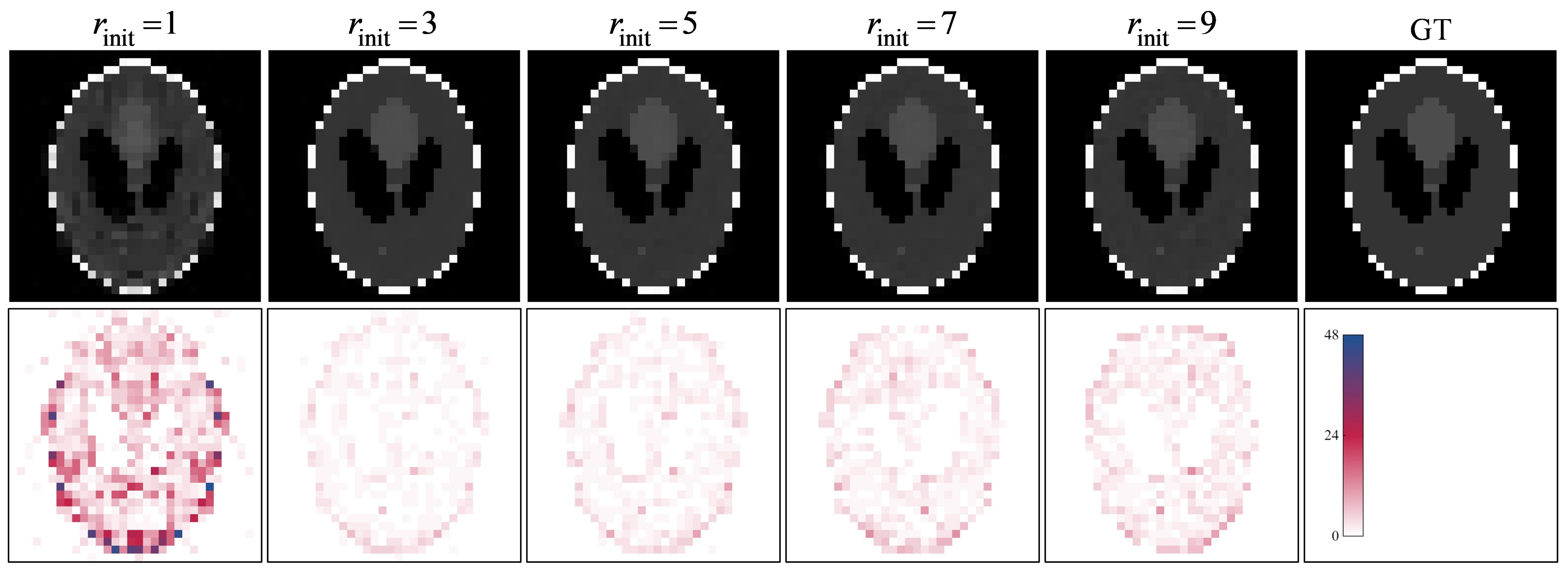}}
\caption{Reconstructions for $r_{\mathrm{init}}\in\{1,3,5,7,9\}$. Each result is paired with its error map; ground truth (GT) is shown in the last column.}
\label{fig:half13579}
\end{figure}

\begin{figure}[!t]
\centerline{\includegraphics[width=1\columnwidth]{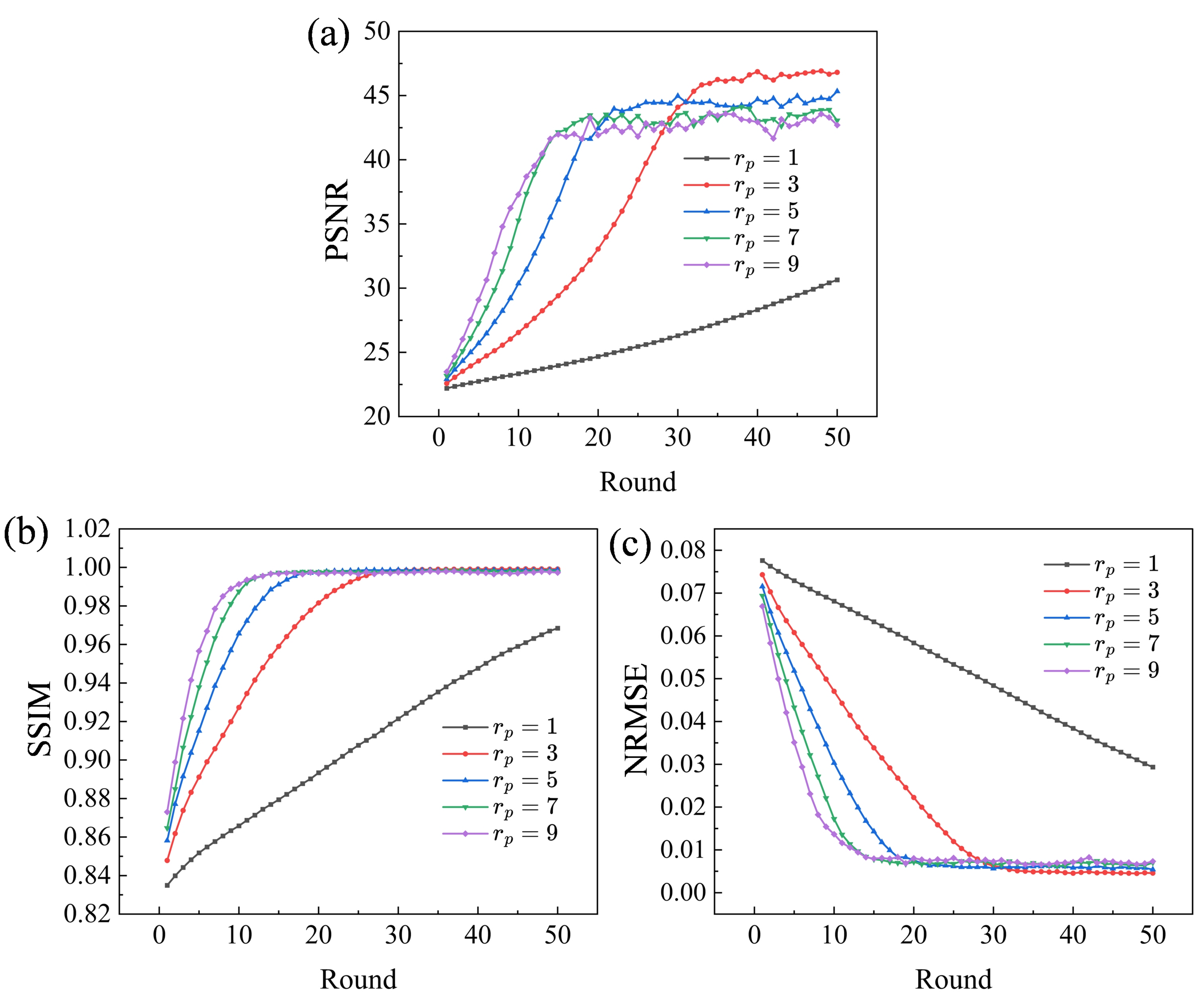}}
\caption{Convergence curves for different $r_{\mathrm{init}}$ values. Intermediate half-widths converge faster and yield better final PSNR, SSIM, and NRMSE.}
\label{fig:halfw_curves}
\end{figure}

 We further studied sensitivity to the initial interval half-width \(r_{\mathrm{init}}\in\{1,3,5,7,9\}\), as shown in Figs.~\ref{fig:half13579} and~\ref{fig:halfw_curves}.

\par When the initial half-width is too small, the search range becomes overly restrictive and cannot adequately correct the current estimate during early refinement. When it is too large, the local search becomes less focused, slowing convergence and weakening fine-scale refinement. Intermediate settings provide a better exploration--refinement trade-off, leading to faster convergence and better final reconstruction quality.

\par This observation is consistent with the main experimental setting, where \(r_{\mathrm{init}}=5\) provides a stable balance between early-stage search flexibility and later-stage local precision under the evaluated proof-of-concept setting.

\subsection{Discussion}

\par The experiments support the feasibility of the proposed framework as a proof of concept for QUBO-based grayscale CT reconstruction with compact binary image encodings. On the evaluated $32 \times 32$ images, dynamic interval encoding refines local intensity intervals instead of relying on fixed global bit planes. Ablations show that the edge-preserving prior stabilizes structures under incomplete data, while objective-term balancing mitigates coefficient-scale mismatch between data-fidelity and prior terms.

\begin{figure}[!t]
\centerline{\includegraphics[width=1\columnwidth]{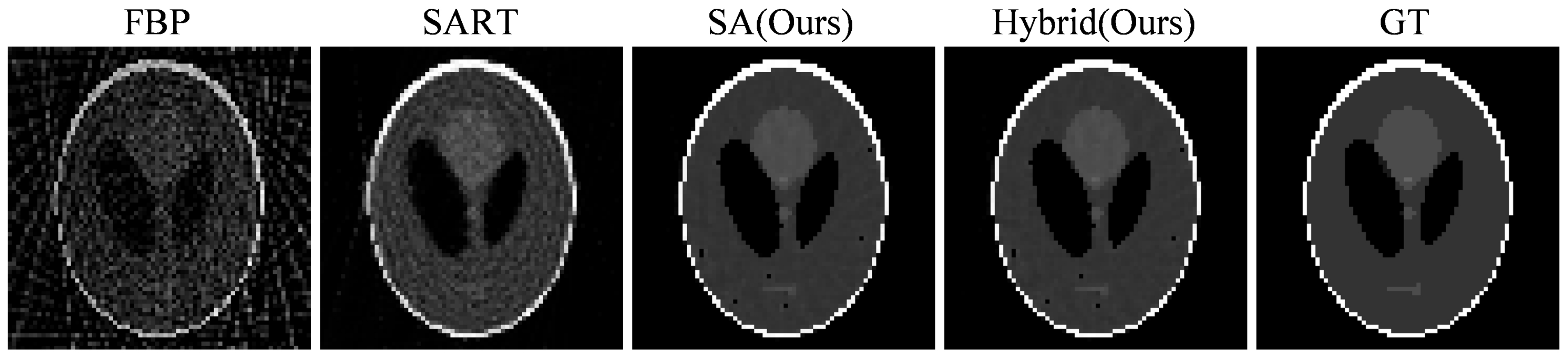}}
\caption{Larger-image sparse-view (20 views) reconstruction on a $72 \times 72$ phantom. FBP, SART, SA-based QUBO reconstruction, D-Wave Hybrid QUBO reconstruction, and ground truth (GT) are shown for visual comparison.}
\label{fig:large72}
\end{figure}

\par Dynamic interval encoding addresses the gray-level bottleneck of fixed global codes. A fixed low-bit code allocates few states over the full gray-value range, causing coarse quantization and structural bias. The proposed local dictionary is updated around the current estimate at each round, keeping each QUBO compact while improving effective gray-level coverage. The boundary-hit rule separates pixels that need broader exploration from those ready for local refinement.

\par We further examined a larger $72 \times 72$ phantom case (Fig.~\ref{fig:large72}), where the active support increases to 2230 pixels and the corresponding QUBO contains 4460 binary variables with approximately $5.42\times 10^6$ quadratic terms under the same 2-bit local encoding. The method still recovers the main phantom structures with both local simulated annealing and the D-Wave hybrid solver, suggesting that the formulation is not restricted to very small images. The best SA and Hybrid runs reach comparable PSNR/SSIM values of 37.81/0.9695 and 37.83/0.9702, respectively, providing a quantitative complement to the visual comparison in Fig.~\ref{fig:large72}. The larger instance also requires longer reconstruction time and remains more difficult to optimize under the current binary-variable and bit-depth constraints, making large-scale QUBO CT reconstruction a key direction for future work.

\section{Conclusion}

\par We presented a QUBO-based quantum CT reconstruction method that combines dynamic interval encoding with prior-balanced optimization. By updating local gray-level codebooks across refinement rounds while keeping each QUBO compact, the method improves grayscale expressivity under a limited binary-variable budget. Experiments on sparse-view and limited-angle fan-beam CT demonstrate improved reconstruction quality over the evaluated baselines and show the effectiveness of the proposed encoding and objective design. These results indicate that representation design is central to quantum-compatible CT reconstruction, where binary encodings must balance gray-level fidelity and problem size. Future work will extend the framework to larger-scale, noisy, and measured CT data, and explore more scalable variable allocation and solver-aware implementations.

\bibliographystyle{IEEEtran}
\bibliography{references}

\end{document}